\documentclass{ametsocV6.1_arxiv}

\newcommand{\arrtime}{{\bm{y}}}
\newcommand{\arrtimeRV}{{\bm{Y}}}

\newcommand{\inputs}{{\bm{x}}}
\newcommand{\inputsRV}{{\bm{X}}}
\newcommand{\inputsRVdist}{{P_{\bm{X}}}}

\newcommand{\cond}{{P_{\bm{Y}|\bm{X}}}}

\newcommand{\joint}{{P_{\bm{Y} \bm{X}}}}
\newcommand{\jointsamp}{{(\bm{Y},\bm{X})}}

\title{Probabilistic Forecasting of Localized Wildfire Spread Based on Conditional Flow Matching}

\authors{Bryan Shaddy,\aff{a}\correspondingauthor{Bryan Shaddy, bshaddy@usc.edu \\ \emph{This Work has not yet been peer-reviewed and is provided by the contributing Author(s) as a means to ensure timely dissemination of scholarly and technical Work on a noncommercial basis. Copyright and all rights therein are maintained by the Author(s) or by other copyright owners. It is understood that all persons copying this information will adhere to the terms and constraints invoked by each Author's copyright. This Work may not be reposted without explicit permission of the copyright owner.}}
Haitong Qin,\aff{b}
Brianna Binder,\aff{a}
James Haley,\aff{c}
Riya Duddalwar,\aff{d}
Kyle Hilburn,\aff{c}
Assad Oberai,\aff{a}
}

\affiliation{\aff{a}{Department of Aerospace and Mechanical Engineering, University of Southern California}\\
\aff{b}{Department of Mathematics, University of Southern California}\\
\aff{c}{Cooperative Institute for Research in the Atmosphere, Colorado State University}\\
\aff{d}{Department of Computer Science, University of Southern California}
}

\abstract{This study presents a probabilistic surrogate model for localized wildfire spread based on a conditional flow matching algorithm. The approach models fire progression as a stochastic process by learning the conditional distribution of fire arrival times given the current fire state along with environmental and atmospheric inputs. Model inputs include current burned area, near-surface wind components, temperature, relative humidity, terrain height, and fuel category information, all defined on a high-resolution spatial grid. The outputs are samples of arrival time within a three-hour time window, conditioned on the input variables. Training data are generated from coupled atmosphere–wildfire spread simulations using WRF-SFIRE, paired with weather fields from the North American Mesoscale model. The proposed framework enables efficient generation of ensembles of arrival times and explicitly represents uncertainty arising from incomplete knowledge of the fire–atmosphere system and unresolved variables. The model supports localized prediction over subdomains, reducing computational cost relative to physics-based simulators while retaining sensitivity to key drivers of fire spread. Model performance is evaluated against WRF-SFIRE simulations for both single-step (3-hour) and recursive multi-step (24-hour) forecasts. Results demonstrate that the method captures variability in fire evolution and produces accurate ensemble predictions. The framework provides a scalable approach for probabilistic wildfire forecasting and offers a pathway for integrating machine learning models with operational fire prediction systems and data assimilation.}

\begin{document}

\maketitle

\section{Introduction}

The increasing impact of wildfires in recent decades has driven the development of a wide range of wildfire spread prediction tools. These tools span a spectrum from simplified two-dimensional models based on semi-empirical rate-of-spread formulations to high-fidelity models that resolve small-scale combustion processes (\cite{bakhshaii2019review,finney1998farsite,linn2002studying,rothermel1972mathematical}). State-of-the-art operational models include coupled atmosphere–wildfire systems such as WRF-SFIRE, which account for fire–weather feedback mechanisms that are critical for accurate prediction of fire behavior (\cite{mandel2011coupled}). More recently, machine learning-based approaches have also been explored for wildfire prediction (\cite{jain2020review}). Despite these advances, even the most sophisticated models exhibit significant deviations from reality over time horizons of one to two days. These limitations arise from the inherently chaotic nature of wildfire spread, incomplete representation of physical processes, and uncertainty in key inputs such as fine-scale fuel structure. This loss of predictive accuracy motivates the use of data assimilation techniques that incorporate observations of wildfire evolution to improve forecasts.

Data assimilation methods for high-dimensional, nonlinear systems typically rely on ensembles of predictions, which poses challenges when using computationally intensive wildfire models. This limitation motivates the development of lightweight yet accurate surrogate models capable of efficiently generating prediction ensembles. While existing physics-based wildfire models capture many of the processes governing fire spread, their computational cost restricts their use in ensemble-based frameworks. These models, however, provide a strong foundation for machine learning-based emulation. By training surrogate models to reproduce the behavior of physics-based simulations, it is possible to achieve substantial computational speedup while retaining key physical characteristics and accuracy.

Existing machine learning approaches to wildfire prediction employ a range of methodologies and data sources. Some approaches resemble cellular automata, where transition rules are learned from historical wildfire data (\cite{zheng2017forest}). Other methods utilize convolutional neural networks trained on historical observations or simulated outputs from models such as FARSITE (\cite{hodges2019wildland,burge2023recurrent,radke2019firecast,bolt2022spatio,marjani2024cnn}). While these approaches demonstrate promise, they often suffer from limitations related to data quality and incomplete representation of relevant physical processes. In many cases, these limitations stem from the use of simplified training data that omit key dynamics, such as fire–weather interactions.

As a result, several existing machine learning-based wildfire prediction models exhibit limited applicability to real-world scenarios, reducing their ability to produce accurate and physically consistent predictions (\cite{andrianarivony2024machine}). Furthermore, several models represent fire progression using coarse temporal metrics, such as burned area or probability of burning, which complicates their integration with data assimilation frameworks that leverage high-resolution observational data from satellites  (\cite{radke2019firecast,khennou2021forest,hodges2019wildland}). In addition, most existing approaches produce deterministic predictions, which limits their utility in ensemble-based settings (\cite{burge2023recurrent,bolt2022spatio}). Constraints on domain size further restrict the applicability of some models to large, long-duration wildfire events.

In this work, we develop a probabilistic surrogate model for wildfire spread based on conditional flow matching (\cite{lipman2022flow, dasgupta2026solving}). The model predicts fire growth over a $3.2~\text{km} \times 3.2~\text{km}$ domain at 25 m resolution, incorporating weather, terrain, and fuel information to simulate fire progression over a 3-hour time step. The proposed approach provides a computationally efficient framework for generating ensembles of fire spread predictions, making it well suited for sequential data assimilation and operational forecasting. The model requires only the current fire extent as input, enabling straightforward initialization from observed perimeters, such as those obtained from airborne measurements, or from previous model outputs. This formulation allows the model to be applied autoregressively to simulate fire evolution over extended time horizons. Additionally, by decomposing the domain into smaller subregions, the approach could be applied to fires of varying sizes while maintaining high spatial resolution. This patch-based strategy also enables targeted prediction of fire growth in regions of interest without requiring simulation of the full fire extent.

The conditional flow matching model is trained on WRF-SFIRE simulations of wildfires occurring across the contiguous United States (CONUS) during 2023, together with corresponding weather data from the North American Mesoscale (NAM) model. Model inputs include the current burned area, zonal (east--west) and meridional (north--south) wind components, relative humidity, temperature, terrain elevation, and fuel category data. The model outputs fire progression over a 3-hour interval in the form of fire arrival times, which indicate the time at which the fire first reaches each location in the domain. This formulation enables incorporation of key drivers of wildfire spread while relying only on readily available atmospheric and static environmental data. The probabilistic framework allows sampling from the conditional distribution of fire arrival times given the inputs, thereby enabling generation of prediction ensembles and quantification of uncertainty arising from incomplete knowledge of factors such as fuel moisture and the fully coupled atmosphere–fire state.

The proposed approach differs from prior methods in several important ways (\cite{hodges2019wildland,burge2023recurrent,radke2019firecast,bolt2022spatio,marjani2024cnn,khennou2021forest}). First, it leverages training data from coupled atmosphere–wildfire simulations of real events, improving generalizability and enabling incorporation of more realistic fire dynamics. Second, it adopts a probabilistic formulation, allowing uncertainty to be explicitly represented and enabling efficient generation of ensembles without requiring input perturbations. Third, the model supports localized prediction of fire spread, allowing simulation of subregions of a fire perimeter without the computational burden associated with full-domain physics-based models.

The remainder of the manuscript is organized as follows. Section~\ref{methods} describes the problem formulation, training data generation, and the conditional flow matching methodology. Section~\ref{results} presents model evaluation, including sensitivity analysis and validation against WRF-SFIRE for both single-step predictions and multi-step recursive forecasts. Finally, Section~\ref{conclusions} summarizes the findings and outlines directions for future work.

\section{Methods} \label{methods}
In the following section we discuss the methods utilized to construct the probabilistic wildfire spread surrogate model developed here. First, we present the formulation for the problem of probabilistic estimation of wildfire spread based on weather, terrain, and fuel data. Next, we describe the construction of training data from WRF-SFIRE solutions and NAM weather data. Finally, we present a description of the flow matching algorithm and how it is trained. 

\subsection{Problem Formulation}

The problem is formulated over a spatial domain of extent $3.2\,\text{km} \times 3.2\,\text{km}$. This domain is discretized using a uniform grid with a resolution of 25 meters, resulting in a computational mesh of $128 \times 128$ points on which all relevant variables are defined.

The primary objective is to predict fire growth over a specified time interval. In particular, given the perimeter of an active fire at time $t_n$, the underlying terrain, the spatial distribution of fuel, and weather conditions evaluated at the intermediate time $t_n + \Delta t/2$, the goal is to determine the evolution of the fire over the interval $(t_n, t_n + \Delta t)$, where $\Delta t = 3$ hours.

Fire growth is represented through an arrival time map, which encodes the time at which the fire reaches each grid point in the domain. This quantity is denoted by $\arrtime$, where $\arrtime \in \Omega_y \equiv \mathbb{R}^{128 \times 128}$, with each entry corresponding to the fire arrival time at a specific spatial location.

The model takes as input the following set of variables:
\begin{itemize}
    \item The burned area at time $t_n$, represented as a single binary field.
    \item Weather variables, including zonal and meridional wind components, temperature, and relative humidity, evaluated at the midpoint between time steps, $t_{n} + \Delta t/2$. These are represented as four real-valued scalar fields.
    \item Static terrain height data, represented as a single real-valued scalar field.
    \item Static fuel category data, in which each Anderson fuel category is encoded as a separate variable using a one-hot representation. This results in fourteen binary scalar fields, including one corresponding to the absence of fuel.
\end{itemize}

Each input variable is defined as a scalar field over the same computational grid as $\arrtime$. All input variables are concatenated to form a composite conditioning tensor, denoted by $\inputs \in \Omega_x \equiv \mathbb{R}^{128 \times 128 \times 20}$.

Although the proposed model incorporates a comprehensive set of input variables, it does not include all variables that define the state in a WRF-SFIRE simulation. For example, omitted variables include those describing fuel moisture and the state of the atmosphere at multiple elevation levels. As a consequence of these omissions, as well as the inherently chaotic nature of wildfire spread, we expect that a given set of input variables may correspond to multiple plausible fire evolution scenarios. This observation motivates the development of a probabilistic model for wildfire spread that explicitly captures this one-to-many mapping. In particular, we represent the predicted fire arrival time as a random vector, $\arrtimeRV \in \Omega_y$, and the input variables as a random vector, $\inputsRV \in \Omega_x$. Our objective is to characterize the conditional density, $\cond$, such that for a given realization $\inputsRV = \inputs$, we can draw samples of $\arrtimeRV$ from this distribution.


To sample from the target conditional density, $\cond$, we employ a conditional generative modeling approach based on conditional flow matching (\cite{lipman2022flow,wildberger2023flow,tong2023conditional,dasgupta2026solving}). The model is trained on samples $\jointsamp$ drawn from the joint distribution $\joint$. Once trained, the flow matching model enables sampling from the learned conditional density $\cond$ for a given realization $\inputsRV = \inputs$, thereby generating corresponding samples of $\arrtimeRV$. These samples can subsequently be used to compute statistics of interest. In particular, we consider the medoid, defined as the sample with the smallest average distance to all other samples, which serves as a representative prediction, as well as the pixel-wise standard deviation, which provides a measure of predictive uncertainty.

Training tuples are constructed by first sampling $\inputsRV$ from the marginal distribution $\inputsRVdist$, which characterizes typical wildfire conditions. For each such sample, a numerical wildfire spread model (WRF-SFIRE in our case) is used to simulate fire progression over a time interval of $\Delta t = 3~\text{hours}$. This yields corresponding samples of $\arrtimeRV$ drawn from the conditional distribution $\cond$. The resulting tuple $\jointsamp$ therefore constitutes a sample from the joint distribution $\joint$. By repeating this procedure, a collection of samples from the joint distribution $\joint$ is obtained and used to train the flow matching model. In practice, this data generation process is carried out by running simulations from ignition and evolving the fire over extended time horizons, thereby producing the required training dataset.

\subsection{Training Data Generation}

We generate training data using the coupled atmosphere–wildfire model WRF-SFIRE, which integrates the Weather Research and Forecasting (WRF) model with a wildfire spread propagation model. The wildfire component, SFIRE, employs a level-set method, with spread rates computed using the semi-empirical Rothermel rate-of-spread model (\cite{mandel2011coupled,rothermel1972mathematical}). We use the \texttt{wrfxpy} system to execute WRF-SFIRE simulations. This system automates the full simulation workflow, including domain configuration, acquisition and processing of meteorological data for initial and boundary conditions, preprocessing of terrain and fuel data, and the execution and monitoring of simulations (\cite{mandel2019interactive}). 

In the following sections, we describe the configuration of the WRF-SFIRE simulations and the procedure used to construct training data from their outputs.


\subsubsection{WRF-SFIRE Solutions}

A total of 140 WRF-SFIRE simulations of wildfires across the contiguous United States (CONUS) for 2023 were generated to construct the training dataset, with an additional 12 simulations reserved for testing and model validation. Incidents were selected based on their simulated spatial extent after an initial 24-hour growth period. Fires that were either too large to be contained within the simulation domain or too small to be representative were manually excluded. Simulations were initialized using ignition locations and times derived from GOES and VIIRS active fire detections provided by NOAA’s Next Generation Fire System (NGFS). The NGFS clusters satellite detections into incident groups and associates them with reported fires cataloged by the National Interagency Fire Center (NIFC). Ignition times were defined as the timestamp of the earliest detection associated with each fire event. Ignition locations were estimated using the average position of all VIIRS detections within the footprint of the initial GOES detection. In cases where no VIIRS detections were available within the first four hours following the initial GOES detection, the ignition location was instead estimated using the average position of the GOES detections.

We conducted the simulations on a domain of size $30~\text{km} \times 30~\text{km}$, with an atmospheric grid resolution of 1 km. The atmospheric model employed 40 vertical levels, with a model top at approximately 5000 Pa (corresponding to an altitude of about 20 km). To represent wildfire dynamics, we used a $1200 \times 1200$ fire grid with a spatial resolution of 25 m, yielding a grid refinement ratio of 40:1 between the atmospheric and fire grids. Ignition locations were positioned near the center of each domain.

We prescribed initial and boundary atmospheric conditions using data from the North American Mesoscale (NAM) Forecast System Grid 227, which has a spatial resolution of 5 km. These data were provided at 3-hour intervals throughout each simulation. Fuel information was obtained from the 2019 LANDFIRE 13 Anderson Fire Behavior Fuel Models (FBFM13) dataset, and terrain elevation data were sourced from the USGS National Elevation Dataset (NED). Using this configuration, we simulated wildfire evolution over the first 48 hours following ignition. These simulations produced fire arrival time maps, which we used to construct both fire area inputs and fire arrival time outputs for training the flow matching model. The corresponding weather inputs were derived from the NAM data used to drive the simulations, while terrain and fuel inputs were obtained from the respective static datasets.

\subsubsection{Training Data Construction}

Using fire arrival time maps obtained from WRF-SFIRE, together with corresponding weather data from NAM at 3-hour intervals, as well as terrain and fuel category data, we construct training data tuples. We use NAM weather data, originally employed to drive the boundary conditions of the WRF-SFIRE simulations, to define the weather inputs for our model. This choice avoids the need to model the fully coupled atmospheric state during prediction and enables the use of readily available forecast data.

We generate training samples through a data augmentation procedure that includes random rotation, forecast time selection, and spatial patch extraction. For each WRF-SFIRE simulation, we denote the full 48-hour fire arrival time map by $\bm{\tau}$ and apply the following steps:

\begin{enumerate}
    \item Resample the weather data onto the 25 m resolution fire grid on which arrival times, terrain, and fuel variables are defined.
    \item Apply a random rotation to $\bm{\tau}$, along with the corresponding weather, terrain, and fuel data, using an angle drawn from the uniform distribution $U(0^{\circ}, 360^{\circ})$.
    \item Rotate the wind vectors accordingly and then project them back onto a fixed east--west and north--south coordinate system.
    \item Sample a forecast time $T_n \sim U(T_{\text{ign}}, T_{\max} - \Delta t)$, where $T_{\text{ign}}$ denotes the ignition time (i.e., the minimum value of $\bm{\tau}$), $T_{\max}$ is the total simulated duration, and $\Delta t = 3$ hours is the prediction interval.
    \item Construct the fire area used as input by assigning all pixels denoted by the index pairs $(i,j)$ a value of $-1$ if $\tau_{ij} \leq T_n$, and a value of 1 if    $\tau_{ij} > T_n$.
    
    \item Construct the fire arrival time to be used as output $y_{ij}$ as follows:
    \begin{equation}
        y_{ij} = \left\{ \begin{array}{ll} T_n, & \tau_{ij} \leq T_n \\
        T_{n} + \Delta t, & \tau_{ij} > T_n + \Delta t \\
        \tau_{ij}, & {\rm otherwise} \end{array} \right.
    \end{equation}
    \item Linearly interpolate between available weather data snapshots to obtain weather variables at time $T_n + \Delta t/2$.
    \item Extract a randomly selected spatial patch of size $128 \times 128$ pixels, ensuring that the patch contains sufficient information from the fire area input.
\end{enumerate}

The sequence of steps described above yields a single sample $\jointsamp$ from a WRF-SFIRE solution. For each WRF-SFIRE solution, we perform 10 rotations described in Step 2 above and for each rotation select 10 forecasting times $T_n$ described in Step 4 above. This results in the generation of 100 samples for each WRF-SFIRE solution, which yields $14,000$ samples for training data and $1,200$ samples for test data. 

The weather, terrain, fuel and output arrival time data are then normalized as follows:
\begin{itemize}
    \item Wind components are normalized by dividing by the standard deviation computed across all samples.
    \item Relative humidity and temperature are normalized by subtracting their respective global mean and dividing by their respective global standard deviation.
    \item Terrain height is first shifted by subtracting the sample-wise minimum, and is then normalized by subtracting the global mean and dividing by the global standard deviation.
    \item Fuel category data are represented using a one-hot encoding, resulting in 14 binary masks that indicate the presence of each fuel type.
    \item Output arrival times are normalized by linearly scaling the $(T_n, T_n + \Delta t)$ interval to $(-1,1)$. 
\end{itemize}

\begin{figure}
    \centering
    \includegraphics[width=0.7\linewidth]{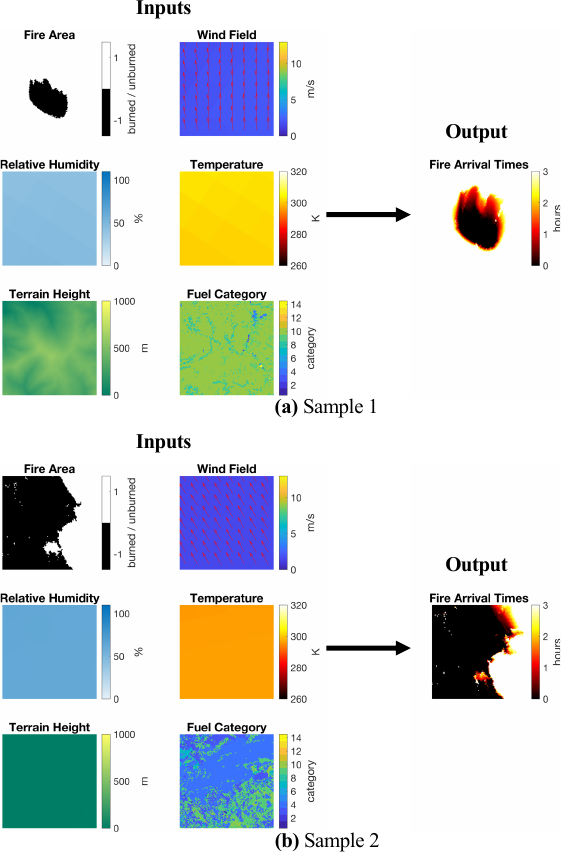}
    \caption{Two training samples, with input variables and corresponding fire arrival time outputs displayed. The wind field shown is plotted using the magnitude and direction computed from the zonal and meridional wind components and fuel category data is additionally displayed as a single categorical image determined from the one-hot encoded data used for training.}
    \label{fig:surrogate_train_data}
\end{figure}

In Figure~\ref{fig:surrogate_train_data} we present two samples from the training data; one where the entire fire extent is contained within the domain and another demonstrating where only a portion of the fire extent is captured in the domain. The training data contains a mixture of these two types of data. 

\subsection{Flow Matching}
In the following section we provide a brief review of the conditional flow-matching algorithm. The reader is referred to \cite{lipman2022flow,wildberger2023flow,tong2023conditional,dasgupta2026solving} for a detailed description of this algorithm. 

We begin by defining a pseudo-time variable $t \in (0,1)$, and for each value of $t$ define a new random variable $\arrtimeRV_t$ defined as 
\begin{equation}
    \arrtimeRV_t = \bm{Z}(1-t) + \arrtimeRV t,
\end{equation}
where $\bm{Z} \sim P_{\bm{Z}} = N(\bm{0},\bm{1})$ is the standard normal random variable, and $\arrtimeRV \sim \cond$. We note that $\arrtimeRV_t$ has a few important properties, including that at time $t=0$, $\arrtimeRV_0 = \bm{Z}$ and at time $t=1$, $\arrtimeRV_1 = \arrtimeRV$. Therefore, at any time $t$, $\arrtimeRV_t$ can be thought of as a mixture of $\bm{Z}$ and $\arrtimeRV$.

As shown in \cite{dasgupta2026solving}, from the definition of $\arrtimeRV_t$, we conclude that there exists a velocity field, $\bm{v}_t(\arrtimeRV,\inputsRV)$ which has the following useful property. If we select  samples $\arrtimeRV_0 \sim P_{\bm{Z}} $, and integrate in pseudo-time using  
\begin{equation} \label{ODE}
    \frac{d\arrtimeRV_t}{dt} = \bm{v}_t(\arrtimeRV_t,\inputsRV),
\end{equation}
then we are guaranteed that the collection $\arrtimeRV_1 \sim \cond$. That is, we can generate samples from the conditional distribution $\cond$ by first generating samples from $P_{\bm{Z}}$ and integrating them in time using the ordinary differential equation above. 

Further, it can be shown that the velocity field used in (\ref{ODE}), denoted by $\bm{v}_t$ is the field that minimizes the loss function,
\begin{equation}
    L(\bm{v}_t) = \int_{0}^{1} \int_{\Omega_y} \int_{\Omega_y} \int_{\Omega_x} |\bm{v}_t(\arrtime_t,\inputs) - (\arrtime-\bm{z})|^2 \joint\jointsamp P_{\bm{Z}}(\bm{z}) d\inputs d\arrtime d\bm{z} dt,
\end{equation}
\noindent where $\arrtime_t = \bm{z}(1-t) + \arrtime t$.

The integral in this loss can be approximated as a Monte Carlo sum by sampling $t^{(i)} \sim U(0,1)$, $\bm{z}^{(i)} \sim N(\bm{0},\bm{1})$, and $(\arrtime^{(i)},\inputs^{(i)}) \sim \joint$, leading to    
\begin{equation} \label{loss}
    L(\bm{v}_t) = \frac{1}{N} \sum_{i=1}^{N} |\bm{v}_{t^{(i)}}(\arrtime_{t^{(i)}},\inputs^{(i)}) - (\arrtime^{(i)} - \bm{z}^{(i)})|^2.
\end{equation}
Note that the evaluation of this loss requires samples $\bm{z}^{(i)}$ from $N(\bm{0},\bm{1})$ and samples of $(\arrtime^{(i)},\inputs^{(i)})$ from $\joint$, which are both available. 

The velocity field in the loss function (\ref{loss}) is approximated using a neural network. Once this network is trained by minimizing the loss, this velocity field can be used in (\ref{ODE}) to obtain a sample of $\arrtime \sim \cond$ for a given input $\inputs$. 

We model the velocity field using an adapted DDPM U-Net architecture, following \cite{ho2020denoising,nichol2021improved}. We implement the network in PyTorch and train it on a single NVIDIA A100 GPU for 45{,}000 mini-batches, corresponding to approximately 320 epochs. We use a batch size of 100 and optimize the model parameters using the AdamW optimizer with a learning rate of $10^{-4}$, weight decay set to 0, $\beta_1 = 0.9$, $\beta_2 = 0.999$, and $\epsilon = 10^{-8}$. Throughout training, we maintain an exponential moving average (EMA) of the model parameters with a decay rate of 0.9999. The final reported model corresponds to these EMA parameters. We compute and track training losses using the EMA parameters, and we select the final checkpoint for evaluation based on the minimum test loss observed during training.

\section{Results} \label{results}
In this section we focus on validating the wildfire spread surrogate model and demonstrating its performance across several scenarios. We begin with a sensitivity analysis, where we evaluate the ability of the surrogate model to reproduce meaningful relations between input weather and terrain variables and the resulting fire spread. Next, we verify the accuracy of the solutions generated by the surrogate model for a single timestep of 3 hours against the corresponding WRF-SFIRE solutions. Lastly, we assess the performance of the model during a recursive roll-out, where fire growth is predicted over a period of 24 hours starting from both an ignition point and from a known burned area for an ongoing wildfire.

\subsection{Sensitivity Analysis}
With the wildfire spread surrogate model trained, we first perform a sensitivity analysis to assess the influence of individual input variables on predicted fire growth. To this end, we prescribe a simplified and controlled set of conditions consisting of spatially uniform weather, terrain, and fuel, together with a circular initial fire perimeter. This configuration removes asymmetries in the wildfire spread problem and enables independent evaluation of the effect of each input variable on the predicted fire growth area. The analysis produces relationships between fire area and each input variable, thereby providing insight into how the model responds to variations in the conditioning inputs. We exclude fuel category from this analysis due to its strong dependence on weather conditions, which would require a more comprehensive, joint sensitivity study across multiple environmental regimes.

We define a control case with uniform wind velocities of $0~\text{m}~\text{s}^{-1}$ in both the zonal ($u$) and meridional ($v$) directions, a relative humidity of $24\%$, a temperature of $80^\circ$F, flat terrain, and a uniform fuel type corresponding to Anderson category 10 (timber, litter, and understory). The initial fire perimeter is additionally specified as a circle with a diameter of $150$ m. Starting from this baseline configuration, we vary each input variable independently to evaluate its impact on the total predicted fire area.

We vary the wind components $u$ and $v$ over the range $[-3, 3]~\text{m}~\text{s}^{-1}$. This range reflects the distribution of wind values in the NAM dataset used for training, where approximately $82\%$ of values lie within this interval. Relative humidity is varied from $0\%$ to $110\%$, and temperature is varied from $40^\circ$F to $115^\circ$F. For terrain, we prescribe an inclined planar surface and vary the angle of inclination from $0^\circ$ to $45^\circ$.

\begin{figure}
    \centering
    \includegraphics[width=.88\linewidth]{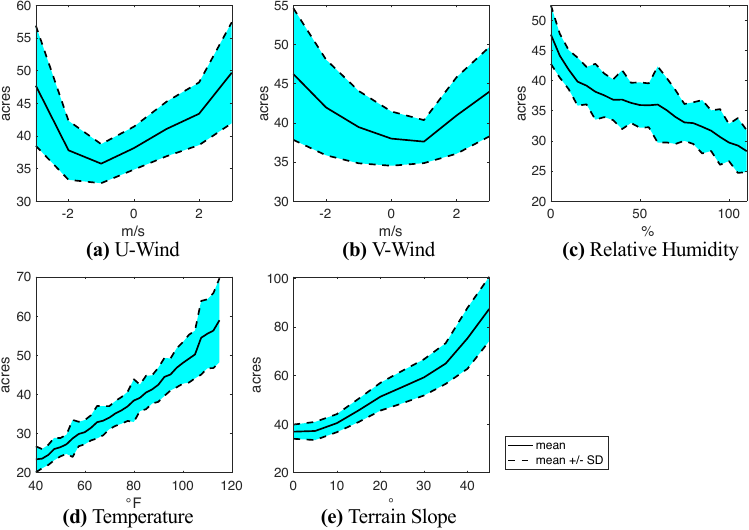}
    \caption{Predicted fire area versus \textbf{(a)} $u$ wind speed, \textbf{(b)} $v$ wind speed, \textbf{(c)} relative humidity, \textbf{(d)} temperature, and \textbf{(e)} terrain slope. For each plot, the mean predicted area is shown along with variability indicated by the standard deviation.}
    \label{fig:surrogate_sensitivity}
\end{figure}

For each set of prescribed conditions, we generate 200 stochastic predictions of fire spread over a single 3-hour interval and compute the corresponding fire growth. From these samples, we estimate the mean and standard deviation of the predicted fire area. We then plot the mean fire area as a function of each input variable and represent variability using the corresponding standard deviation. Figure~\ref{fig:surrogate_sensitivity} summarizes the results of this sensitivity analysis.

The results in Figure~\ref{fig:surrogate_sensitivity} indicate that the model captures physically meaningful relationships consistent with the training data. As wind magnitude increases, the model predicts increased fire growth, with minimal growth occurring near $0~\text{m}~\text{s}^{-1}$. A slight asymmetry is observed in the wind sensitivity curves, with the minimum shifted marginally away from zero.

For relative humidity, the model predicts decreasing fire growth with increasing humidity, consistent with the expected dampening effect of atmospheric moisture. Similarly, higher temperatures lead to increased predicted fire growth. Finally, as the terrain slope increases, the model predicts larger fire areas, reflecting the known influence of slope on fire spread dynamics. Together, these results demonstrate that the surrogate model reproduces key qualitative dependencies observed in wildfire behavior.

\subsection{Validation for three hour interval}
\begin{figure}
    \centering
    \includegraphics[width=.98\linewidth]{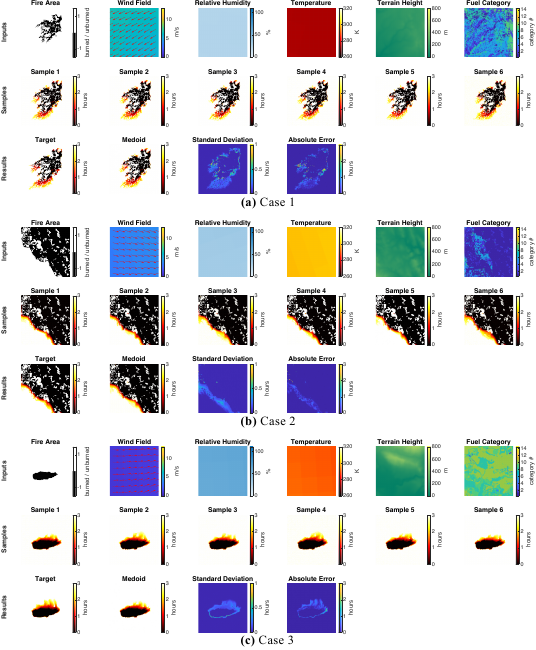}
    \caption{Results for three cases from the testing dataset. The first row shows conditioning inputs, the second row shows six sampled predictions, and the third row presents the target, medoid prediction, pixel-wise standard deviation, and absolute error between the medoid and target.}
    \label{fig:single_timestep}
\end{figure}

Having demonstrated that the model captures expected physics-based relationships, we next evaluate its performance on realistic wildfire scenarios drawn from the testing dataset, which includes heterogeneous and complex weather, terrain, and fuel conditions. We apply the model to 1{,}200 test cases, generating 200 stochastic predictions of fire progression over a single 3-hour interval for each case. From these samples, we compute the medoid prediction, which serves as a representative estimate, as well as the pixel-wise standard deviation, which quantifies predictive uncertainty. Figure~\ref{fig:single_timestep} presents results for three representative cases. This figure includes the input conditions, a subset of generated samples, the medoid prediction, the associated standard deviation plot, the target fire progression, and the absolute error between the medoid and the target.

The results in Figure~\ref{fig:single_timestep} indicate that the model generates a set of plausible fire spread scenarios that are consistent with the input conditions and closely approximate the target solutions. The generated samples exhibit no extreme outliers or non-physical behavior. In all cases, the medoid prediction captures the overall pattern of fire growth, although localized discrepancies in arrival times are observed. For example, in the third case, the medoid slightly under-predicts fire spread along the northern boundary. However, examination of the full set of generated samples reveals that some realizations more closely match the target, indicating that the predictive distribution often encompasses the true outcome. Notably, regions of higher prediction error tend to coincide with regions of elevated standard deviation, suggesting that the model appropriately reflects increased uncertainty in the areas where the error is higher.

\begin{figure}
    \centering
    \includegraphics[width=\linewidth]{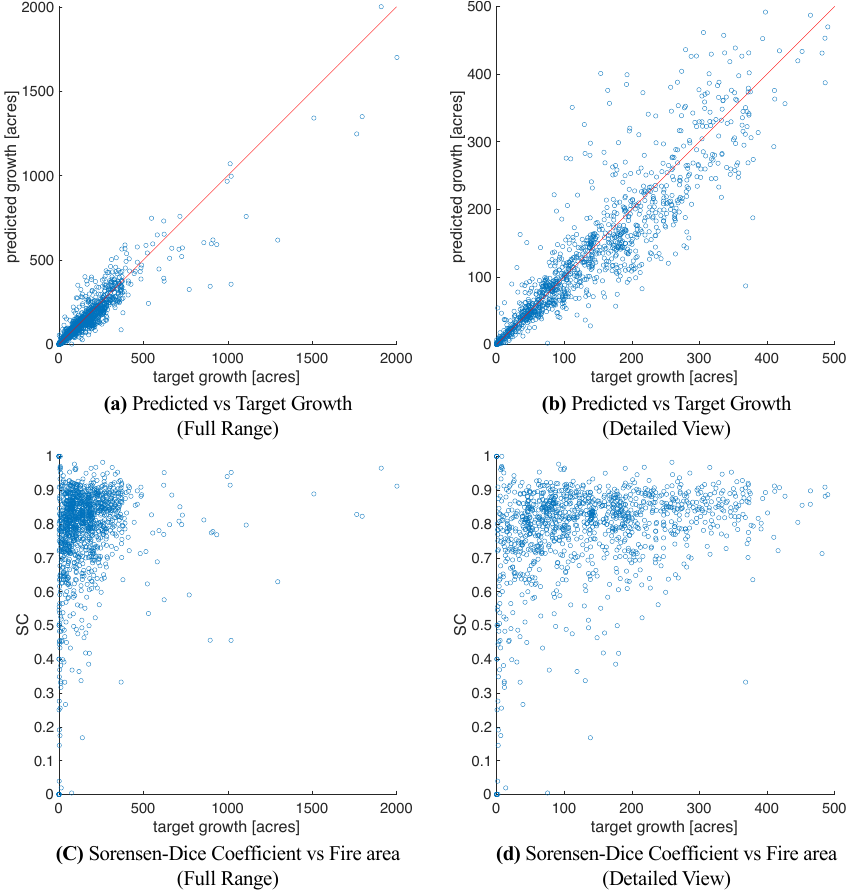}
    \caption{Performance of the wildfire spread surrogate model on the testing dataset. Panels \textbf{(a)} and \textbf{(b)} show medoid predicted versus target burned area over the full and restricted ranges, respectively. Panels \textbf{(c)} and \textbf{(d)} show the Sørensen--Dice coefficient (SC) as a function of target burned area over the same ranges.}
    \label{fig:SC_vs_area}
\end{figure}

To further quantify performance across all 1{,}200 test cases, we compare the total predicted fire growth with the corresponding target values. For each case, we compute the number of grid cells burned over the 3-hour interval in both the medoid prediction and the target. Panels \textbf{(a)} and \textbf{(b)} of Figure~\ref{fig:SC_vs_area} show scatter plots of predicted versus target burned area over the full range of values and over a restricted range for smaller fires. In most cases, the predicted growth closely matches the target, with points clustering near the line of perfect agreement. Some spread is observed, which is expected given the reduced set of input variables available to the surrogate model compared to the full WRF-SFIRE simulations.

We further assess spatial agreement between predicted and target burned areas using the Sørensen--Dice coefficient (SC). For each test case, we compute the number of true positive pixels (TP), false positives (FP), and false negatives (FN), where TP denotes pixels correctly predicted as burned, FP denotes pixels predicted as burned but not burned in the target, and FN denotes pixels burned in the target but not predicted as such. The SC is then given by
\begin{equation}
    SC = \frac{2TP}{2TP + FP + FN}.
\end{equation}
Panels \textbf{(c)} and \textbf{(d)} of Figure~\ref{fig:SC_vs_area} present SC values as a function of target burned area. Most cases yield SC values between 0.8 and 0.9, indicating strong spatial agreement. Lower SC values are more common for smaller fires, where even minor spatial discrepancies can significantly affect the metric. Nevertheless, as seen in Figure~\ref{fig:SC_vs_area}\textbf{(a)}, the model generally captures the correct magnitude of fire growth even in these cases.

\subsection{Recursive Prediction from Ignition}
We next evaluate the recursive application of the surrogate model to predict wildfire growth over multiple timesteps. In this setting, we initialize the model with a known fire area, either from an ignition point or an observed perimeter. The model then predicts fire growth over a single timestep using the corresponding weather inputs. The predicted fire arrival times are subsequently converted into a binary burned/unburned map, which serves as input for the next time step along with updated weather data. This procedure is repeated for the desired number of timesteps.

To propagate uncertainty through time, we adopt an ensemble-based approach. Starting from a single initial fire area, we generate an ensemble of predictions for the first timestep. For subsequent timesteps, each ensemble member produces a single prediction, resulting in a one-to-one mapping between ensemble members and predictions. In this way, the ensemble size remains constant throughout the recursive process.

\begin{figure}
    \centering
    \includegraphics[width=0.74\linewidth]{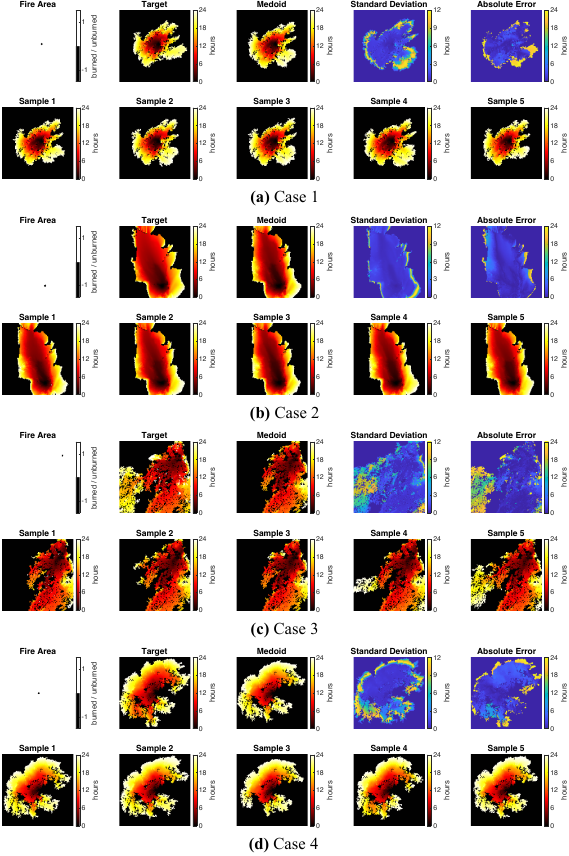}
    \caption{Results from recursive application of the wildfire spread surrogate model to predict growth from ignition to 24 hours for four cases. The first row shows the initial fire area, target growth, medoid prediction, standard deviation, and absolute error. The second row shows representative samples of predicted fire growth.}
    \label{fig:recursive_prediction_from_ignition}
\end{figure}

We apply the surrogate model recursively to predict wildfire spread from ignition up to 24 hours for 12 test cases drawn from the WRF-SFIRE simulations in the testing dataset. These cases were not used during training. For each case, we generate 200 realizations of fire arrival times, requiring approximately nine minutes per case on a single NVIDIA A100 GPU. Figure~\ref{fig:recursive_prediction_from_ignition} presents results for four representative cases, including the initial fire area, the target 24-hour fire arrival time map, the medoid prediction, the pixel-wise standard deviation, and the absolute error between the medoid and the target.

The results in Figure~\ref{fig:recursive_prediction_from_ignition} indicate that the model produces reasonable multi-step growth predictions, with the distribution of generated samples capturing the target evolution in most cases. Some discrepancies are observed in the medoid predictions. For example, in case 3, a region of growth on the western side of the fire is under-predicted by the medoid. However, other ensemble members capture this growth more accurately, and the corresponding region exhibits elevated standard deviation, indicating increased uncertainty.

Across all 12 cases, the model achieves an average Sørensen--Dice coefficient of 0.87 when comparing predicted (medoid) and target burned areas after 24 hours, indicating strong spatial agreement. We further evaluate performance using the probability of detection (POD) and false alarm ratio (FAR), defined in terms of true positives (TP), false positives (FP), and false negatives (FN) as
\begin{equation}
    POD = \frac{TP}{TP + FN}, \quad FAR = \frac{FP}{TP + FP}.
\end{equation}
The POD ranges from 0 to 1, with higher values indicating better detection, while the FAR ranges from 0 to 1, with lower values indicating fewer false positives. Across the 12 cases, we obtain an average POD of 0.85 and an average FAR of 0.07, demonstrating that the model captures fire growth effectively without substantial over- or under-prediction.

Finally, we assess the relationship between predictive uncertainty and error by computing the Pearson correlation coefficient between the pixel-wise standard deviation and the pixel-wise absolute error across all cases. The average correlation of 0.46 indicates a moderate positive relationship, suggesting that regions of higher predicted uncertainty tend to correspond to larger prediction errors. This result highlights the utility of the uncertainty estimates provided by the model.

\subsection{Recursive Prediction from a Known Perimeter}
We next apply the surrogate model to localized subsections of actively burning fires in order to predict growth from a known perimeter. As in the ignition-based setting, we begin with a single initial fire area and generate an ensemble of predictions for the first timestep. Subsequent predictions proceed in a one-to-one manner, where each ensemble member produces a single realization at each timestep. 

\begin{figure}
    \centering
    \includegraphics[width=0.74\linewidth]{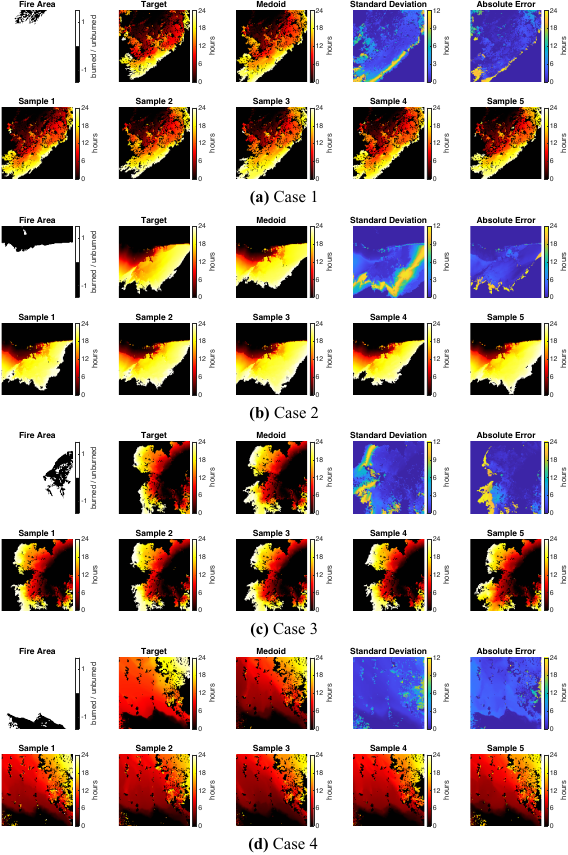}
    \caption{Results from recursive application of the wildfire spread surrogate model to a subsection of an active fire, initialized from the 12-hour perimeter and propagated to 36 hours. The first row shows the initial fire area, target growth, medoid prediction, standard deviation, and absolute error. The second row shows representative prediction samples.}
    \label{fig:recursive_prediction_from_perimeter}
\end{figure}

We apply the model to the same 12 wildfire cases used in prior evaluations, now initializing from the fire extent at 12 hours after ignition. The computational domain is positioned along a portion of the fire perimeter. From this initial condition, we generate 200 realizations of fire growth over a 24-hour period. Figure~\ref{fig:recursive_prediction_from_perimeter} presents results for four representative cases. Plots for each case include the initial fire extent, the target spread solution, the medoid prediction, the pixel-wise standard deviation, and the absolute error between the medoid and the target, along with representative samples.

The results in Figure~\ref{fig:recursive_prediction_from_perimeter} show that the medoid prediction agrees well with the target spread solution across all cases. As in the predictions that were initialized from ignition, the generated samples exhibit a reasonable diversity of plausible fire spread scenarios, with the standard deviation providing a meaningful measure of variability. The highest variability is typically observed along the advancing fire front, indicating that uncertainty is primarily associated with the extent of fire growth over the prediction horizon. For example, in case 1, elevated uncertainty is observed near the southern boundary of the fire, which coincides with regions of larger prediction error. This further highlights the utility of the uncertainty estimates.

Across the 12 cases, the model achieves an average Sørensen--Dice coefficient of 0.90, a probability of detection (POD) of 0.96, and a false alarm ratio (FAR) of 0.14 for 24-hour growth predictions initialized from a 12-hour perimeter. Relative to predictions initialized from ignition, these results indicate a modest improvement in SC and POD, accompanied by a slight increase in FAR, suggesting a tendency toward mild over-prediction in this setting. The correlation between predicted uncertainty and error remains moderate, with an average Pearson correlation coefficient of 0.45.

Overall, these results demonstrate that the surrogate model performs effectively when applied to subsections of actively burning fires, while providing substantial computational speed-up relative to physics-based models. In contrast, physics-based approaches face challenges in initializing from partial fire perimeters and typically require simulation of the entire fire domain, making localized prediction significantly more computationally demanding.

\section{Conclusions and Future Work} \label{conclusions}

In this work, we have developed a probabilistic surrogate model for wildfire spread that enables the efficient generation of ensembles of fire progression predictions. The model is based on conditional flow matching and is trained using simulations of historical wildfires generated with the coupled atmosphere–wildfire model WRF-SFIRE, together with corresponding weather data from the North American Mesoscale (NAM) model. Given the current fire extent, along with wind, relative humidity, temperature, terrain, and fuel information, the model predicts fire progression over a 3-hour interval. The model relies exclusively on readily available inputs and does not require the execution of additional parameterizations or atmospheric models. Furthermore, it can be applied to localized subsections of a fire perimeter, enabling targeted predictions without simulating the full fire extent. Because the model requires only the current fire area as input, it can additionally be initialized using observed perimeters, such as those obtained from airborne measurements, without requiring knowledge of prior fire evolution.

We evaluated the model through a sensitivity analysis designed to assess the learned relationships between input variables and predicted fire growth. The results demonstrated that the model captures physically meaningful dependencies, with fire growth responding appropriately to variations in weather and terrain conditions. 

We further evaluated its performance on 1{,}200 test cases, corresponding to 3-hour (single timestep) fire growth. In each case, we used the model to generate 200 samples. These results showed that the model produces realistic fire spread patterns over a single time step and achieves strong agreement with target solutions. Using the Sørensen--Dice coefficient (SC) to quantify spatial agreement, we found that most cases yielded values between 0.8 and 0.9. Lower SC values were observed for smaller fires, reflecting the sensitivity of the metric to small spatial discrepancies. Nevertheless, the model accurately captured the overall magnitude of fire growth in these cases, indicating that it reliably represents the dominant fire dynamics.

To assess longer-term predictive performance, we applied the model recursively to simulate 24 hours of fire growth starting from ignition. This procedure involved eight sequential predictions, with each step using the output of the previous step as input. Across 12 test cases, the model achieved an average Sørensen--Dice coefficient (SC) of 0.87, a probability of detection (POD) of 0.85, and a false alarm ratio (FAR) of 0.07, indicating strong agreement with the reference WRF-SFIRE simulations and limited over-prediction. The Pearson correlation coefficient between predicted uncertainty (standard deviation) and error was 0.46, demonstrating that the model provides informative uncertainty estimates that are moderately correlated with error. 

We also evaluated the model on subsections of active fires by initializing from a 12-hour perimeter and predicting 24 hours of growth. In this setting, the model achieved an average SC of 0.90, POD of 0.96, and FAR of 0.14 across 12 cases indicating strong overall performance. The correlation between uncertainty and error remained consistent, with an average value of 0.45.

The proposed surrogate model provides a substantial computational speed-up relative to physics-based approaches, generating 200 samples for a single 3-hour prediction in just over one minute while maintaining reliable accuracy. These characteristics make the model well suited for applications such as sequential data assimilation, where rapid generation of prediction ensembles is essential. The model also shows promise for operational deployment, offering probabilistic forecasts of wildfire spread along with associated uncertainty estimates that can support decision-making in firefighting and resource allocation.

Despite these advances, several challenges remain. In its current form, the model can be applied independently to subsections of large fires; however, improved domain decomposition strategies are needed to ensure consistency and information exchange across adjacent regions, thereby avoiding discontinuities in predicted fire growth (\cite{chan1994domain, dolean2015introduction}). Future work will focus on integrating the surrogate model with data assimilation frameworks, including coupling with update models based on conditional generative approaches (\cite{shaddy2024generative,shaddy2026generative}), to enable real-time assimilation of observational data and further enhance predictive capability.

\clearpage
\acknowledgments

The NOAA Bipartisan Infrastructure Law project NA22OAR4050672I supported the creation of the forecasts. The NASA Disasters project 80NSSC19K1091 supported the high-performance computing for running the forecasts. The NASA FireSense project 80NSSC23K1344 supported development and evaluation of the machine learning approach. AAO acknowledges support from ARO grant W911NF2410401. The authors acknowledge the Center for Advanced Research Computing (CARC) at the University of Southern California, USA for providing the computing resources that have contributed to the research results reported within this publication.

\bibliographystyle{ametsocV6}
\bibliography{references}

\end{document}